\documentclass[conference]{IEEEtran}
\IEEEoverridecommandlockouts
\usepackage{cite}
\usepackage{amsmath,amssymb,amsfonts}
\usepackage{algorithmic}
\usepackage{graphicx}
\usepackage{textcomp}
\usepackage{xcolor}
\usepackage{booktabs}
\usepackage[inline]{enumitem}
\def\BibTeX{{\rm B\kern-.05em{\sc i\kern-.025em b}\kern-.08em
    T\kern-.1667em\lower.7ex\hbox{E}\kern-.125emX}}

\begin{document}

\title{Reflective VLM Planning for Dual-Arm Desktop Cleaning: Bridging Open-Vocabulary Perception and Precise Manipulation}

\author{Yufan Liu\textsuperscript{\dag*}, Yi Wu\textsuperscript{\dag*}, Gweneth Ge\textsuperscript{\dag}, Haoliang Cheng\textsuperscript{\ddag}, and Rui Liu\textsuperscript{\S}\\
\textsuperscript{\dag}Robotics Institute, Carnegie Mellon University\\
\textsuperscript{\ddag}Department of Electrical and Computer Engineering, Carnegie Mellon University\\
\textsuperscript{\S}College of Aeronautics \& Engineering, Kent State University\\
\textsuperscript{*}Equal contribution
}

\maketitle
\begin{abstract}
Desktop cleaning demands open-vocabulary recognition and precise manipulation for heterogeneous debris. We propose a hierarchical framework integrating reflective Vision-Language Model (VLM) planning with dual-arm execution via structured scene representation. Grounded-SAM2 facilitates open-vocabulary detection, while a memory-augmented VLM generates, critiques, and revises manipulation sequences. These sequences are converted into parametric trajectories for five primitives executed by coordinated Franka arms. Evaluated in simulated scenarios, our system achieving 87.2\% task completion, a 28.8\% improvement over static VLM and 36.2\% over single-arm baselines. Structured memory integration proves crucial for robust, generalizable manipulation while maintaining real-time control performance.
\end{abstract}


\section{Introduction}

Robotic desktop cleaning presents a compelling intersection of open-vocabulary perception and precise manipulation. Unlike structured industrial tasks, domestic environments necessitate recognition of arbitrary items, distinguishing valuables from debris, and coordinated multi-scale manipulation—from fine-grained grasping to broader sweeping motions.

Traditional modular approaches often struggle with semantic ambiguities in such settings. While Vision-Language Models (VLMs) offer potent open-vocabulary understanding\cite{ren2024groundedsamassemblingopenworld, feng2025reflectiveplanningvisionlanguagemodels,ahn2022icanisay}, bridging the semantic-geometric gap for precise, real-time robotic control remains a key challenge.

Our hierarchical architecture separates strategic reasoning from tactical execution via structured information flow. We posit that effective VLM integration requires structured intermediate representations over end-to-end language generation. Our system employs Grounded-SAM2 for detection and a memory-augmented VLM on scene graphs to generate JSON plans, converted to trajectories by a lightweight interface.

Our primary technical contributions are: (1) a structured VLM interface creating geometrically precise manipulation plans from RGB-D observations and instance masks; (2) a reflective planning mechanism using execution feedback via rolling memory buffers for robust dynamic plan revision; and (3) a dual-arm coordination framework enabling simultaneous, collision-avoidant object manipulation and debris sweeping.

Evaluated across 17 simulated desktop scenarios featuring diverse objects and debris, our reflective dual-arm system achieved 87.2\% task completion (vs. 58.4\% for a static VLM baseline) with improved efficiency. These findings highlight structured VLM integration with reflection as a viable path toward generalizable domestic robotics.

\begin{figure}[t]
    \centering
    \includegraphics[width=0.9\linewidth]{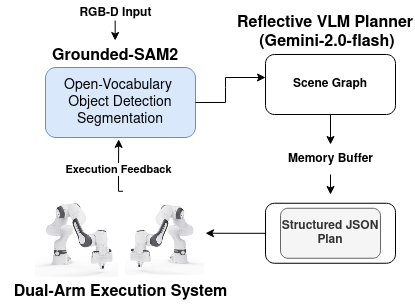} 
    \caption{Conceptual overview of our hierarchical system. RGB-D input is processed by Grounded-SAM2, generating a scene graph. The reflective VLM planner uses this graph and memory to produce a JSON task plan, which is then executed by the dual-arm system. Execution feedback closes the loop.}
    \label{fig:pipeline}
\end{figure}
\section{Method}
Our approach addresses desktop cleaning through a three-layer architecture (Fig.~\ref{fig:pipeline}) that bridges open-vocabulary perception with precise dual-arm manipulation. A key principle is the use of structured intermediate representations to facilitate effective VLM integration while preserving the geometric precision crucial for manipulation.

\subsection{Perception-to-Planning Pipeline}

\paragraph{Structured Scene Understanding} RGB-D observations are processed by Grounded-SAM2, yielding instance masks, bounding boxes, and semantic labels for detected objects. Subsequently, a scene graph is constructed, integrating object properties (identity, 3D pose, size, category) with spatial relationships (e.g., containment, support, proximity) derived from further point cloud analysis. This structured representation preserves essential geometric information while facilitating subsequent natural language-based reasoning.

\paragraph{VLM Interface Design} The constructed scene graph is serialized into a textual description, which, along with task specifications and execution history, forms the input to the VLM (Gemini-2.0-flash). The VLM generates manipulation plans as structured JSON objects according to a predefined schema. This schema defines sequences of five parameterized primitives: \textsc{Pick}, \textsc{Place}, \textsc{Consolidate}, \textsc{Wipe}, and \textsc{Inspect}. Parameters for each primitive include target object(s), desired poses, force thresholds, and post-execution verification flags.

\subsection{Reflective Planning with Memory}
Adaptive behavior is achieved via a rolling memory buffer storing the five most recent plans and their execution outcomes. Upon primitive failure—detected through force feedback, pose errors, or timeouts—a replanning cycle is triggered. During replanning, the VLM receives updated visual information (current bounding boxes and masks), select previous images, the history of prior plans and outcomes (memory context), and a structured prompt. This prompt guides the VLM to reflect on the encountered failure and revise its strategy. This reflective loop enables dynamic plan revision while maintaining consistency, closing the reasoning-level perception-action cycle in approximately 250\,ms, while real-time servo control remains uninterrupted.

This memory mechanism directly addresses a critical limitation of static VLM approaches: their inability to recover from execution failures. By embedding execution history into subsequent prompts, the system can learn from its mistakes iteratively without requiring external supervisory signals for adaptation.

\subsection{Dual-Arm Coordination}
Coordinated bimanual manipulation is pivotal for efficiently performing simultaneous object consolidation and debris sweeping. The dual Franka arms operate under a unified impedance controller, with real-time collision avoidance facilitated by a shared occupancy map and model-predictive trajectory modification. All five manipulation primitives support single-arm or coordinated execution, while \textsc{Consolidate} and \textsc{Wipe} are specifically designed for synergistic bimanual actions.

This hierarchical architecture effectively balances high-level semantic reasoning with the stringent low-level geometric constraints of manipulation, thereby enabling generalizable robotic capabilities without compromising control precision or safety.

\section{Experiments}
\paragraph{Simulation Environment} We evaluate our system in Isaac Sim 4.5 across 17 distinct scenarios featuring diverse debris types, object arrangements, and variations in desktop height. Each scenario populates a $0.9 \times 0.6$\,m desktop surface with 6--15 rigid objects (e.g., cups, books, writing implements), 3--5 fragile items (e.g., eyeglasses, small electronic components), and at least two types of loose debris (e.g., 1--3\,mm granular crumbs, 15--20\,mm paper shreds).

\paragraph{Task Definition} A trial is deemed successful if the robot achieves three objectives: (1) all valuable objects are relocated to a designated safe tray; (2) all debris is swept into a specified collection zone; and (3) no fragile items are damaged or improperly displaced. Each trial has a maximum duration of 180\,s or concludes earlier upon successful task completion or unrecoverable errors.

\subsection{Results}

\begin{table}[t]
\centering
\caption{Performance comparison across 17 scenes (3 seeds each). Our reflective dual-arm system achieves highest success rates with improved efficiency.}
\label{tab:main}
\begin{tabular}{@{}lcccc@{}}
\toprule
Method & Success (\%) ↑ & Time (s) ↓ &  Recovery ↑ \\
\midrule
\textbf{Ours (Reflective + Dual)} & \textbf{87.2} & \textbf{118} & \textbf{0.82} \\
Static VLM & 58.4 & 139 & 0.00 \\
Single Arm & 51.0 & 164 & 0.69 \\
\bottomrule
\end{tabular}
\end{table}

\paragraph{Reflection Enables Recovery.} Removing memory (Static VLM) reduces success by 28.8 percentage points and eliminates failure recovery, confirming that adaptive replanning is essential for robust manipulation.

\paragraph{Dual-Arm Coordination Matters.} Single-arm operation reduces success by 36.2 percentage points and increases episode time by 28\%, primarily because sweeping and pick-place operations must be serialized rather than parallelized.

\section{Conclusion}

We presented a hierarchical framework for dual-arm desktop cleaning that combines open-vocabulary perception with reflective VLM planning. Our system bridges the semantic-geometric gap through structured scene representations while enabling adaptive behavior through memory-augmented reflection. Experimental evaluation demonstrates significant improvements over static baselines, confirming that structured VLM integration provides a promising pathway toward generalizable domestic manipulation. The work establishes foundations for future investigations into tactile-enhanced manipulation and real-time VLM deployment in robotic systems.

\bibliographystyle{IEEEtran}
\bibliography{references}

\end{document}